\documentclass[lettersize,journal]{IEEEtran}
\usepackage{amsmath,amsfonts}
\usepackage{algorithmic}
\usepackage{algorithm}
\usepackage{array}
\usepackage[caption=false,font=normalsize,labelfont=sf,textfont=sf]{subfig}
\usepackage{textcomp}
\usepackage{stfloats}
\usepackage{url}
\usepackage{xcolor}
\usepackage{verbatim}
\usepackage{graphicx}
\usepackage{cite}
\usepackage{booktabs}
\usepackage{multirow}
\usepackage{pifont} 
\newcommand{\cmark}{\ding{51}}
\newcommand{\xmark}{\ding{55}}
\usepackage[colorlinks=true]{hyperref}
\usepackage{amsmath,amsfonts}
\usepackage[ruled,vlined,linesnumbered,algo2e]{algorithm2e}
\usepackage{array}
\usepackage[caption=false,font=normalsize,labelfont=sf,textfont=sf]{subfig}
\usepackage{textcomp}
\usepackage{stfloats}
\usepackage{url}
\usepackage{verbatim}
\usepackage{graphicx}
\usepackage{xcolor}
\usepackage{cite}
\usepackage[utf8]{inputenc}
\usepackage{booktabs} 
\usepackage{multirow} 
\usepackage{graphicx} 
\usepackage{amsmath}  
\usepackage{amsthm}
\newtheorem{definition}{Definition}
\usepackage[table]{xcolor} 

\begin{document}

\title{Visually Irrelevant Hallucinations in Medical Vision-Language Models}
%
\author{Ke Zou, Nan Zhou, Tao Yang, Meng Liu, Linchao He, Zehui Liao, Ling Huang, Guangren Huang, Yidi Chen,\\
Yi Zhang, \IEEEmembership{Senior Member, IEEE}, Hu Chen, \IEEEmembership{Member, IEEE},
and Huazhu Fu, \IEEEmembership{Senior Member, IEEE}
\thanks{This work was supported by the Sichuan Provincial Natural Science Foundation under Grant 2026NSFSC0426. }
\thanks{K. Zou is with the Department of Ophthalmology, Yong Loo Lin School of Medicine, National University of Singapore, Singapore 119228.}\thanks{N. Zhou, T. Yang, M. Liu, L. He, and H. Chen are with the College of Computer Science, Sichuan University, Chengdu 610065, China.}
\thanks{Z. Liao is with the Faculty of Information Technology, Monash University, Clayton, Victoria 3800, Australia. }
\thanks{L. Huang is with the Institute of Clinical Sciences, Imperial College London, London W12 0NN, UK.}
\thanks{G. Huang is with the Department of Radiology, Beihai People's Hospital, Beihai, Guangxi 536000, China. Y. Chen is with the Department of Radiology, The First Affiliated Hospital of Guangxi Medical University, Nanning, Guangxi 530021, China.}
\thanks{Y. Zhang is with the School of Cyber Science and Engineering and the Key Laboratory of Data Protection and Intelligent Management, Ministry of Education, Sichuan University, Chengdu 610065, China.}
\thanks{H. Fu is with the institute of Advanced Intelligence and Computing, Agency for Science, Technology and Research (A*STAR), Singapore 138632. }
\thanks{Ke Zou and Nan Zhou are contributed equally to this work.}
\thanks{Hu Chen and Huazhu Fu are the co-corresponding authors (e-mail: huchen@scu.edu.cn, hzfu@ieee.org). }
}

\markboth{Journal of \LaTeX\ Class Files,~Vol.~14, No.~8, August~2021}%
{Shell \MakeLowercase{\textit{et al.}}: A Sample Article Using IEEEtran.cls for IEEE Journals}


\maketitle

\begin{abstract}
Medical vision--language models (VLMs) have shown strong capabilities in radiology report generation (RRG) and medical visual question answering (Med-VQA), generating free-form clinical descriptions and answers from medical images. However, these models may generate clinically meaningful statements that are irrelevant to the visual evidence present in the input image. Such statements may lack sufficient visual grounding or contradict the observed image content. We refer to such errors as \emph{visually irrelevant hallucinations} and investigate them at the phrase level. To detect these hallucinations, we propose CLEAR (\textbf{C}ounterfactual \textbf{L}ocalized \textbf{E}valuation of \textbf{A}lignment and \textbf{R}esponse), a model-agnostic post-hoc framework for phrase-level hallucination risk estimation. CLEAR first decomposes the generated response into sentence-level, visually assessable clinical phrases. For each phrase, phrase-conditioned grounding identifies a candidate visual region as an auditing target, which is selectively perturbed through localized counterfactual intervention. CLEAR then characterizes the intervention using two complementary auditing cues: \emph{response cues}, capturing semantic and clinically meaningful changes in the generated response, and \emph{alignment cues}, measuring changes in image--phrase compatibility. These cues are normalized and jointly modeled to estimate phrase-level hallucination risk. Across four datasets spanning RRG and Med-VQA, CLEAR consistently outperforms state-of-the-art hallucination detection baselines across multiple medical and general-purpose VLMs. For example, on MIMIC-CXR, CLEAR improves AUPRC from 85.10\% to 90.23\%. Its effectiveness is further validated on a GT-derived controlled VIH benchmark. These results establish visually irrelevant hallucination as an important fine-grained reliability failure mode in medical VLMs and demonstrate the value of model-agnostic visual-support auditing for more trustworthy multimodal clinical generation. Code is  available at \url{https://github.com/Zhounan1222/VIH}

\end{abstract}

\section{Introduction}
\label{sec:introduction}

Recent advances in vision--language models (VLMs) have enabled
substantial progress in multimodal clinical applications, particularly
radiology report generation (RRG) and medical visual question answering
(Med-VQA)
~\cite{zhang2023biomedclip,li2023llava,moor2023foundation,zhang2024biomedgpt,qin2026lmod,li2026advancing,sellergren2025medgemma,
xu2025lingshu,yuan2023ramm}.
Despite their strong generative capabilities, factual reliability
remains a major barrier to trustworthy clinical deployment
~\cite{cao2025boosting,zhang2024vision}.
A clinically plausible response may contain generated statements that
lack sufficient support from, or are inconsistent with, the input
image, even when they are expressed with high linguistic fluency and
apparent confidence
~\cite{chen2024detecting,zhang2024radflag,
yan2025medhalltune,ostmeier2024green}.

\begin{figure}
    \centering
    \includegraphics[width=1\linewidth]{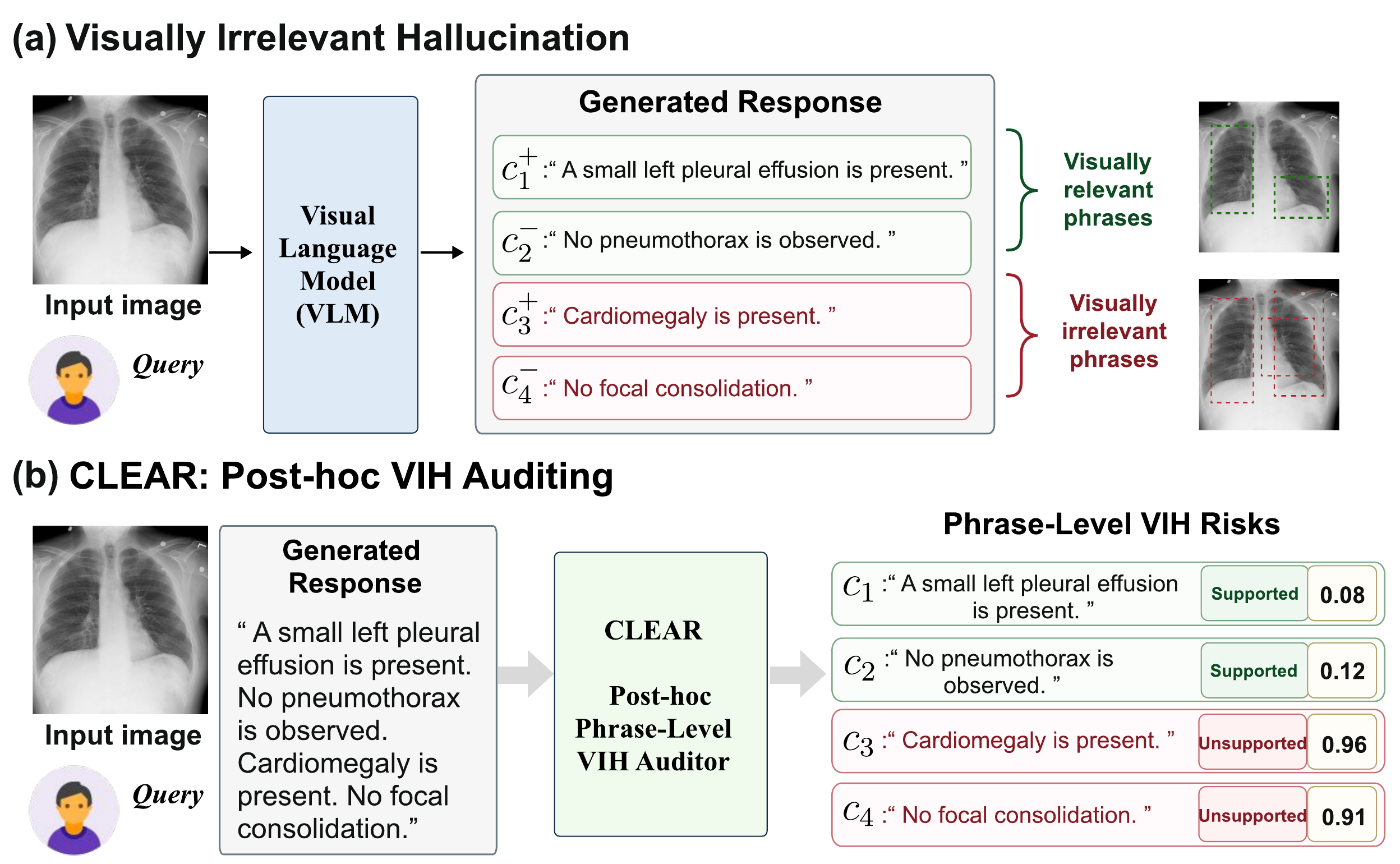}
    \caption{
Motivation and overview of phrase-level visually irrelevant hallucination
(VIH) detection.
(a) A fluent VLM response may contain both visually supported and
irrelevant clinical phrases, including positive and negative statements;
therefore, response-level assessment cannot localize unreliable content.
(b) Given the image $I$, query $Q$, generated response $R$, CLEAR performs post-hoc auditing and
estimates a VIH risk  for each atomic, visually assessable
phrase $S_i$. The same formulation applies to both RRG and Med-VQA.
}
\label{fig:motivation}
\end{figure}

Hallucinations in VLMs have traditionally been characterized at the
object, attribute, and relational levels
~\cite{liu2024survey,rani2024visual,
sahoo2024comprehensive,wu2026evaluating}. In this work, we introduce a complementary visual-relevance perspective, termed \emph{visually irrelevant hallucination} (VIH). We define VIH as generated clinical content that is irrelevant to the visual evidence present in the input image, including content that is insufficiently grounded in or inconsistent with the observed image findings. As illustrated in Fig.~\ref{fig:motivation}, visually relevant and irrelevant clinical content may coexist within the same response, making response-level reliability scores insufficient for identifying localized hallucinations.

Existing hallucination detection methods exploit uncertainty,
generation consistency, visual grounding, or external verification \cite{chen2024detecting,xiao2025flair,
farquhar2024detecting,liao2025vision,
cohen2023lm,zhang2024radflag,
an2022attention,hashmi2024envisioning,
yu2024hallucidoctor,sahu2024pelican,
yin2024woodpecker}.
However, none of these signals provides a definitive measure of visual
support in isolation. Hallucinated phrases can remain confident and
stable across repeated generations
~\cite{favero2024multimodal,leng2024mitigating,
farquhar2024detecting}, while a grounding model may localize a
plausible anatomical region even when that region does not substantiate
the corresponding phrase
~\cite{xiao2024can,huy2025seeing,zou2025uncertainty}.
Conversely, valid negative, diffuse, or globally distributed findings
may exhibit only weak changes under localized perturbation. These
observations reveal a fundamental ambiguity in using any single proxy
to determine whether a generated clinical phrase is visually supported.

Accordingly, we formulate VIH detection as a post-hoc fine-grained visual-relevance auditing problem and operationalize it at the phrase level. Rather than treating grounding as evidence of factuality, we use it only to identify a \emph{candidate visual region} whose relationship to each clinical phrase can be probed through localized counterfactual intervention. This enables joint assessment of intervention-induced changes in the generated response and image--phrase alignment.

To operationalize this formulation, we propose CLEAR (\textbf{C}ounterfactual \textbf{L}ocalized \textbf{E}valuation of \textbf{A}lignment and \textbf{R}esponse), a model-agnostic post-hoc framework for phrase-level VIH detection. CLEAR first decomposes the generated response into atomic, visually assessable clinical phrases. For each phrase, phrase-conditioned grounding identifies a candidate visual region as an auditing target, which is selectively perturbed through localized counterfactual intervention. Two complementary auditing cues are used to characterize each localized intervention: \emph{response cues} capture semantic and clinically meaningful changes in the generated response, whereas \emph{alignment cues} quantify changes in image--phrase compatibility. The resulting cues are normalized and jointly modeled to estimate phrase-level hallucination risk.

We evaluate CLEAR across RRG and Med-VQA using multiple medical and general-purpose VLMs. In addition to evaluation on naturally generated responses from public datasets, we construct a GT-derived controlled VIH benchmark, with hallucination labels adjudicated by consensus between two radiologists, to provide transparent and reproducible assessment of visually relevant and irrelevant phrases. The main contributions of this work are fourfold:
\begin{itemize}

\item \emph{VIH} is introduced and formalized as a distinct reliability failure mode in medical VLMs, together with a phrase-level formulation for assessing the visual relevance of generated clinical content.

\item \emph{CLEAR} is proposed as a model-agnostic post-hoc framework for fine-grained VIH detection without requiring access to the internal parameters, gradients, or intermediate representations of the evaluated VLM.

\item \emph{Localized counterfactual auditing} is developed for probing phrase--visual-evidence dependence through joint characterization of intervention-induced changes in generated responses and image--phrase alignment.

\item A comprehensive evaluation protocol is established across four public RRG and Med-VQA datasets and a GT-derived controlled VIH benchmark. Extensive experiments across diverse medical and general-purpose VLMs demonstrate consistent improvements over state-of-the-art hallucination detection baselines.

\end{itemize}

\section{Related Work}
\label{sec:related_work}

\subsection{Hallucination Detection in VLMs}
Hallucination in VLMs broadly refers to generated content that is not adequately grounded in, or is inconsistent with, the visual input. Such errors may arise at the object, attribute, and relation levels, reflecting insufficient visual grounding, reliance on language priors, or failures in multimodal reasoning~\cite{liu2024survey,rani2024visual,xu2024lvlm,wu2026evaluating,sahoo2024comprehensive}. In medical applications, hallucinations may manifest as false clinical findings, incorrect anatomical or laterality descriptions, mischaracterized disease severity, or diagnostic statements that conflict with the corresponding image~\cite{chen2024detecting,yan2025medhalltune,ostmeier2024green}. Importantly, such statements may remain clinically plausible and linguistically fluent despite being poorly related to the visual evidence present in the image.

Existing hallucination detection and verification methods can be organized according to their primary sources of reliability evidence, although these categories are not mutually exclusive. Learned detectors identify hallucinated content using manually annotated or synthetically corrupted supervision~\cite{xiao2025flair,gunjal2024detecting,whitehead2024pre}. Model-based verification methods employ cross-examination, auxiliary VLMs, clinical evaluators, or self-correction procedures to assess generated responses~\cite{cohen2023lm,zhang2024radflag,yu2024hallucidoctor,sahu2024pelican,yin2024woodpecker}. Uncertainty- and consistency-based methods instead analyze token probabilities, predictive entropy, semantic agreement, or repeated generations~\cite{farquhar2024detecting,liao2025vision,li2024reference,zhang2024vl}, while visually grounded approaches examine image--text alignment, localized attribution, or model behavior under modified visual inputs~\cite{an2022attention,hashmi2024envisioning,hu2022x,raghavan2024attention,leng2024mitigating,wang2024mitigating}. In the medical domain, MediHall Score~\cite{chen2024detecting} characterizes clinically relevant hallucination types, MedHallTune~\cite{yan2025medhalltune} studies instruction-based medical hallucination evaluation and mitigation, and GREEN~\cite{ostmeier2024green} identifies structured errors in reports, including false findings, anatomical misidentification, severity errors, and comparison discrepancies.

Although these methods provide meaningful signals of factuality and reliability, they generally do not explicitly assess whether individual clinical phrases within a generated response are relevant to the visual evidence present in the current image. A response may appear clinically plausible or factually compatible with prior knowledge while still containing specific phrases that lack a meaningful relationship to the observed image content. Moreover, response-level correctness, consistency, or verifier scores provide limited insight into which particular clinical phrases are visually justified and which are not. This gap motivates VIH, which focuses on the fine-grained relationship between generated clinical content and the visual evidence from which that content is expected to arise.

\subsection{Visual Grounding and Perturbation-Based Verification}
Visual grounding and attribution methods associate textual concepts with spatial image regions. Attention visualization, gradient-based saliency, and explainable image--text retrieval have been used to identify regions associated with model predictions~\cite{an2022attention,hashmi2024envisioning,hu2022x,raghavan2024attention}. In radiology, grounded generation models such as MAIRA-2~\cite{bannur2024maira} generate findings together with corresponding anatomical regions, while uMedGround~\cite{zou2025uncertainty} jointly extracts diagnostic phrases from free-text reports and localizes them through uncertainty-aware grounding. However, grounding primarily establishes spatial association and does not by itself determine whether a generated phrase is visually relevant or factual. An anatomically plausible region may still be identified for an irrelevant statement, while subtle or diffuse findings may not admit a precise localized correspondence. Thus, for VIH detection, the key question is not simply \emph{where} a phrase can be grounded, but whether its generation actually depends on the associated visual evidence.

Visual perturbation provides a complementary mechanism for probing such dependence. VCD~\cite{leng2024mitigating} and related contrastive decoding methods~\cite{wang2024mitigating} modify the visual input to expose reliance on visual information or language priors, while self-introspective approaches use alternative visual conditions to improve factual generation~\cite{huo2024self}. CounterVHD~\cite{song2026counterfactual} further examines whether predicted grounding changes under counterfactual entity perturbations. These methods demonstrate the utility of counterfactual visual manipulation, but commonly operate at the image or entity level, focus on generation correction or grounding consistency, or rely on a single behavioral signal. They do not explicitly combine localized intervention, response changes, and image--phrase alignment to estimate the visual relevance of a generated clinical phrase.

Our prior work, CoEV~\cite{zhou2026hallucination}, introduced localized counter-evidence verification for positive clinical findings using a deterministic four-quadrant rule. However, CoEV was restricted to positive phrases and treated predefined counterfactual response patterns as direct decision criteria, leaving negative and normal statements, heterogeneous intervention responses, and continuous phrase-level risk insufficiently modeled. CLEAR generalizes this direction in three respects. First, it performs polarity-agnostic auditing across positive, negative, and normal clinical phrases. Second, grounding is used only to identify a \emph{candidate visual region}, rather than being interpreted as evidence of factuality. Third, localized intervention is treated as a probe whose effects are jointly characterized through response cues and image--phrase alignment cues and mapped to phrase-level VIH risk by an evidence-conditioned estimator. In this way, CLEAR assesses whether generated clinical content is visually relevant to the image without relying on localization, perturbation behavior, or linguistic consistency in isolation.

\begin{figure*}
    \centering
    \includegraphics[width=1\linewidth]{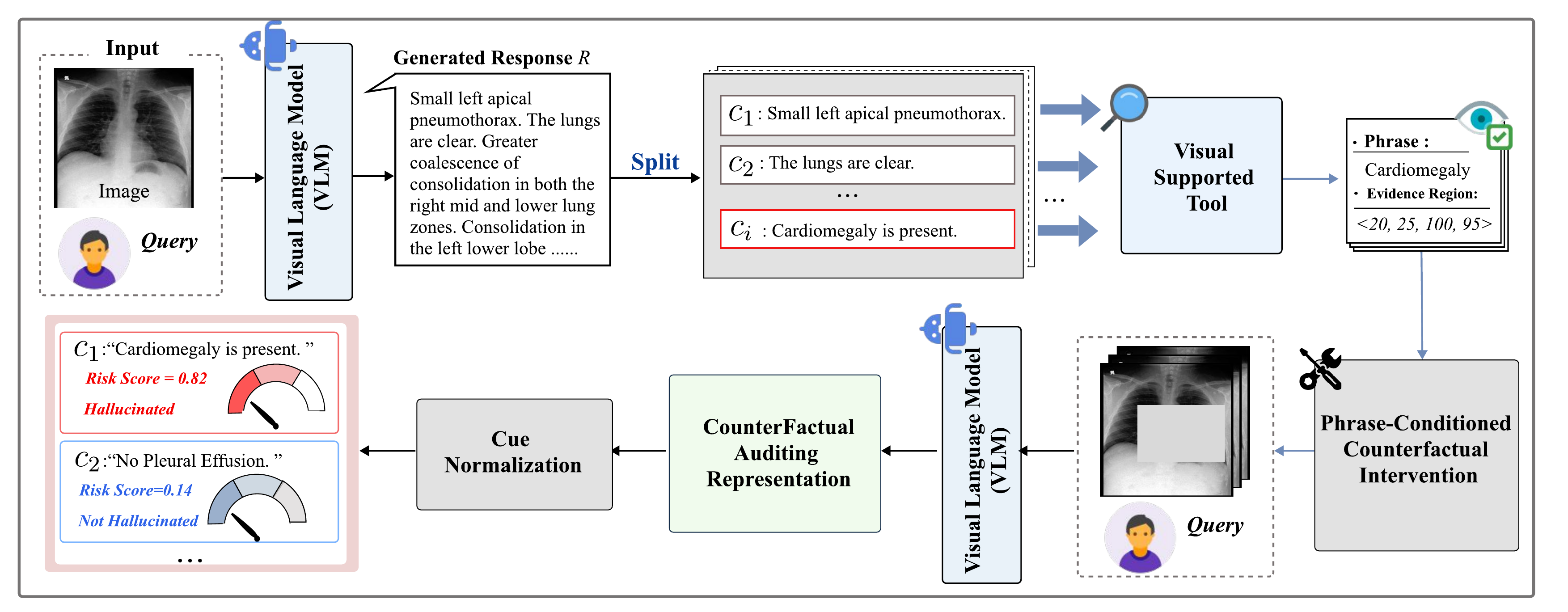}
   \caption{
Overview of CLEAR. The VLM-generated response is decomposed into clinical phrases. For each phrase, a candidate visual region is localized and perturbed, after which the same VLM generates a counterfactual response.
}

    \label{fig:framework}
\end{figure*}

\section{Method}
\label{sec:method}

\subsection{Problem Formulation and Framework Overview}
\label{sec:problem_formulation}

Let $\mathcal{I}$, $\mathcal{Q}$, and $\mathcal{R}$ denote the
spaces of medical images, textual queries or instructions, and
free-form textual responses, respectively. We consider a family of
queryable vision--language models
\begin{equation}
\mathcal{F}
=
\left\{
f
\;\middle|\;
f:\mathcal{I}\times\mathcal{Q}\rightarrow\mathcal{R}
\right\},
\label{eq:vlm_family}
\end{equation}
where each $f\in\mathcal{F}$ denotes an arbitrary VLM that can be
queried with an image and a textual instruction. CLEAR is
model-agnostic and does not require access to the internal parameters,
gradients, or intermediate representations of $f$.

Given a VLM $f\in\mathcal{F}$, an input image
$I\in\mathcal{I}$, and a textual query or instruction
$Q\in\mathcal{Q}$, the model generates
\begin{equation}
R=f(I,Q),
\label{eq:original_generation}
\end{equation}
where $Q$ may correspond to a question in medical visual question
answering or an instruction for medical report generation.

\begin{definition}[Visually Irrelevant Hallucination]
Given a medical image $I$ and a VLM-generated response $R$, a
\emph{visually irrelevant hallucination} (VIH) refers to explicitly
generated and visually assessable clinical content that lacks a meaningful relationship to the visual evidence present in \(I\) under the given task context, including content that is insufficiently grounded in or inconsistent with the observed image content. This definition concerns the visual relevance of explicitly generated clinical content and excludes omitted findings and content that cannot be
adjudicated from the visual input alone.
\label{def:VIH}
\end{definition}

To operationalize VIH detection at a fine-grained level, CLEAR
decomposes the generated response into atomic, visually assessable
clinical phrases using a phrase decomposition function $\mathcal{S}$:
\begin{equation}
\mathcal{S}(R)
=
\left\{
c_i
\right\}_{i=1}^{N_R},
\qquad
N_R
=
\left|
\mathcal{S}(R)
\right|,
\label{eq:phrase_decomposition}
\end{equation}
where $c_i$ denotes the $i$-th atomic, visually assessable clinical phrase explicitly generated in $R$, and $N_R$ denotes the total number of decomposed phrases. A \emph{phrase-level VIH} is defined as a phrase $c_i$ that satisfies Definition~\ref{def:VIH}. Each phrase $c_i$ is associated with a binary VIH label
$Y_i\in\{0,1\}$:
\begin{equation}
Y_i =
\begin{cases}
0, & I \models_v c_i,\\
1, & I \not\models_v c_i.
\end{cases}
\label{eq:hallucination_label}
\end{equation}
Accordingly, $I \models_v c_i$ denotes that $c_i$ is visually relevant to the evidence present in $I$ under task context $Q$, whereas $I \not\models_v c_i$ denotes a phrase-level VIH.

The objective of phrase-level VIH detection is to estimate
\begin{equation}
h_i^{*}
=
P
\left(
Y_i=1
\mid
I,Q,R,c_i
\right),
\label{eq:conditional_hallucination_risk}
\end{equation}
where $h_i^{*}$ denotes the underlying VIH risk associated with
phrase $c_i$. Accordingly, a model-agnostic post-hoc detector
$\mathcal{H}_{\phi}$ is constructed to predict
\begin{equation}
\widehat{\mathbf h}
=
\mathcal{H}_{\phi}(I,Q,R;f)
=
\left\{
\widehat h_i
\right\}_{i=1}^{N_R},
\label{eq:detection_objective}
\end{equation}
where $\widehat h_i\in[0,1]$ estimates $h_i^{*}$. A larger
$\widehat h_i$ indicates a higher estimated risk that the clinical
content expressed by $c_i$ is visually irrelevant. Importantly, CLEAR treats the grounding output as a \emph{candidate visual region} rather than definitive evidence of visual relevance or factuality. For each
phrase, localized counterfactual intervention is applied to the
candidate region, and the resulting changes in the generated response
and image--phrase alignment are jointly analyzed to estimate phrase-level
VIH risk.

\subsection{Phrase-Conditioned Counterfactual Intervention}
\label{sec:counterfactual_intervention}
For each phrase $c_i\in\mathcal{S}(R)$, CLEAR first identifies its
candidate visual region using a phrase-conditioned grounding function:
\begin{equation}
\mathcal{B}_i
=
\mathcal{G}(I,c_i)
=
\left\{
\mathbf b_{i,k}
\right\}_{k=1}^{K_i},
\label{eq:phrase_grounding}
\end{equation}
where $\mathcal{B}_i$ denotes the candidate visual region or
collection of regions associated with phrase $c_i$, and $K_i$ denotes
the number of localized bounding boxes. Importantly, $\mathcal{B}_i$ is treated as an auditing target rather than definitive evidence that $c_i$ is visually relevant to the input image, since the grounding model may produce incomplete or inaccurate localizations. If no valid grounding box is identified, $\mathcal{B}_i$ is defined as the full image region.

Given $\mathcal{B}_i$, a localized counterfactual intervention
operator $\mathcal{T}$ constructs
\begin{equation}
I_i'
=
\mathcal{T}
\left(
I,\mathcal{B}_i
\right),
\label{eq:counterfactual_image}
\end{equation}
where $I_i'$ denotes the counterfactual image associated with phrase
$c_i$. The intervention selectively disrupts the visual content
within the candidate visual region while preserving the
remaining image context as much as possible.

The counterfactual image is subsequently re-queried using the same
VLM and textual instruction:
\begin{equation}
R_i'
=
f
\left(
I_i',Q
\right),
\label{eq:counterfactual_generation}
\end{equation}
where $R_i'$ denotes the counterfactual response corresponding to
phrase $c_i$.

Equations~\eqref{eq:original_generation} and
\eqref{eq:counterfactual_generation} use the same VLM and textual
query; only the visual input is changed. Comparing $R$ with $R_i'$
therefore allows CLEAR to examine how the generated clinical content
responds to localized disruption of candidate visual region. No
reference report or ground-truth answer is required during this process.

\subsection{Counterfactual Auditing Representation}
\label{sec:counterfactual_auditing}

The localized intervention introduced in
Section~\ref{sec:counterfactual_intervention} produces paired
observations $(I,R)$ and $(I_i',R_i')$. Rather than relying on any
single proxy, CLEAR summarizes the effect of the intervention using
two complementary counterfactual auditing cues.

For each phrase $c_i$, the counterfactual auditing representation is
defined as
\begin{equation}
\mathbf z_i
=
\operatorname{Concat}
\left(
\mathbf z_i^{r},
\mathbf z_i^{a}
\right),
\label{eq:auditing_representation}
\end{equation}
where $\mathbf z_i^{r}$ and $\mathbf z_i^{a}$ denote the
counterfactual response and image--phrase alignment cues,
respectively. No individual cue is regarded as definitive evidence
that a phrase is visually relevant or visually irrelevant. Instead,
the representation preserves complementary intervention-induced
observations that are subsequently interpreted jointly by the VIH
risk estimator.

\subsubsection{Counterfactual Response Cues}
\label{sec:response_cues}

Counterfactual response cues characterize how the generated clinical content changes after localized disruption of the candidate region. The counterfactual response $R_i'$ is first decomposed using the same phrase decomposition function $\mathcal{S}$ defined in Eq.~\eqref{eq:phrase_decomposition}.

\begin{figure}
    \centering
    \includegraphics[width=1\linewidth]{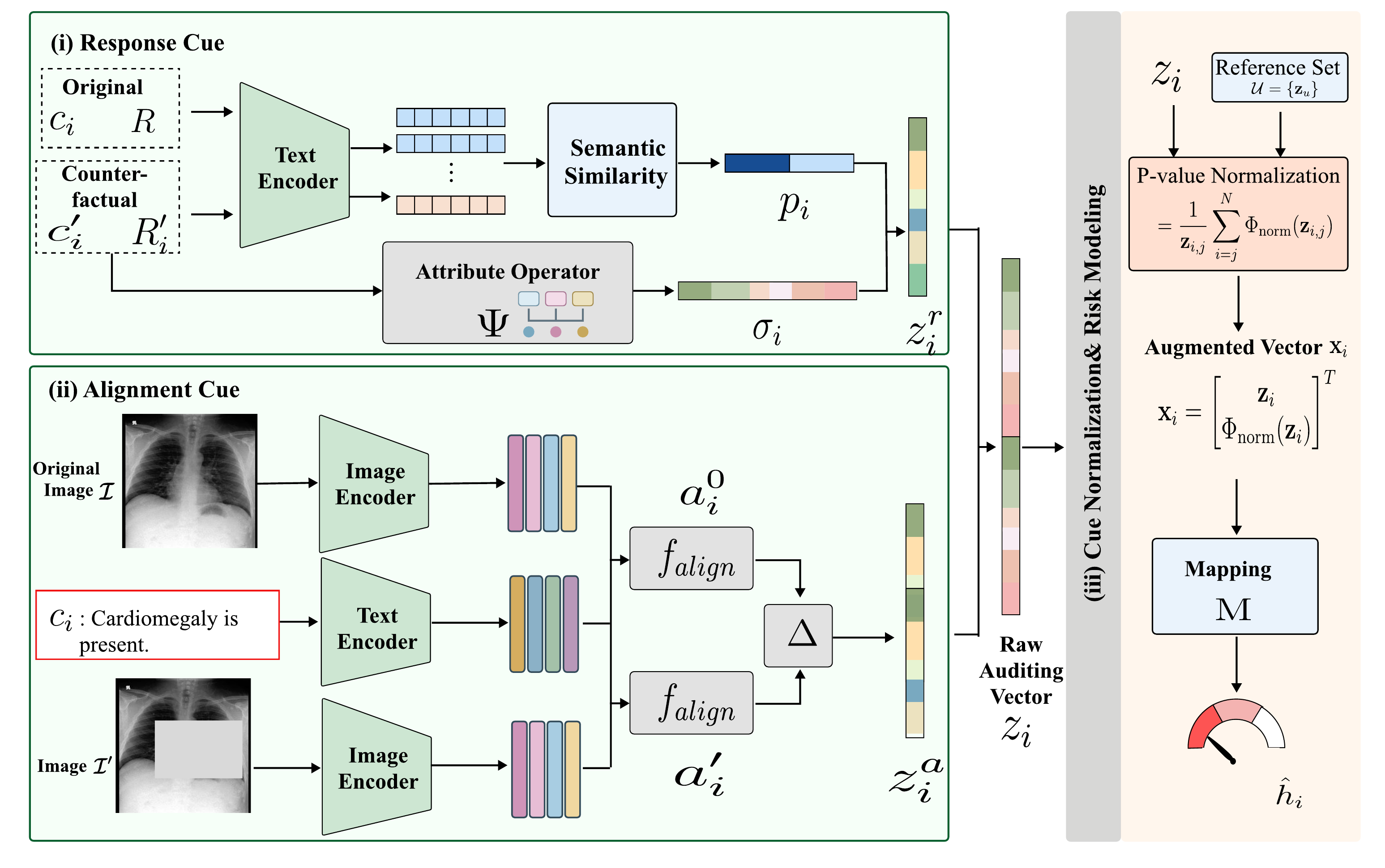}
   \caption{
Architectural overview of Counterfactual Auditing Representation and Calibration in CLEAR.
}

    \label{fig:framework}
\end{figure}

For the original phrase $c_i$, its most semantically corresponding
phrase in $R_i'$ is identified as
\begin{equation}
\bar{c}_i'
=
\underset{c'\in\mathcal{S}(R_i')}{\arg\max}
\;
\operatorname{sim}_{t}
\left(
c_i,c'
\right),
\label{eq:counterfactual_phrase_matching}
\end{equation}
where $\operatorname{sim}_{t}(\cdot,\cdot)$ denotes a textual
semantic similarity function.

The semantic persistence of $c_i$ under intervention is then defined
as
\begin{equation}
p_i
=
\operatorname{sim}_{t}
\left(
c_i,\bar{c}_i'
\right)
=
\max_{c'\in\mathcal{S}(R_i')}
\operatorname{sim}_{t}
\left(
c_i,c'
\right).
\label{eq:semantic_persistence}
\end{equation}
A larger $p_i$ indicates that the semantic content of $c_i$ remains
more persistent after its candidate visual region has been disrupted.


Semantic persistence alone may miss clinically meaningful changes after
intervention. We therefore compute a lightweight linguistic response
descriptor
\begin{equation}
\boldsymbol{\delta}_i
=
\Psi_{\mathrm{r}}
\left(c_i, R, \{R_{i,m}'\}_{m=1}^{M}\right),
\label{eq:clinical_response_descriptor}
\end{equation}
where $c_i$ is the target claim, $R$ is the original response, and
$\{R_{i,m}'\}_{m=1}^{M}$ are the counterfactual responses from different
interventions. In our implementation, $\Psi_{\mathrm{r}}$ extracts simple
clinical linguistic cues, including negation change, uncertainty, entity
retention, entity overlap with the original response, and whether the claim is
an affirmative clinical assertion. These cues are used together with semantic
and visual response features in the final risk estimator.

The counterfactual response cue is represented as
\begin{equation}
\mathbf z_i^{\mathrm{r}}
=
\left[
p_i;
\boldsymbol{\delta}_i
\right],
\label{eq:response_cues}
\end{equation}
where $[\cdot;\cdot]$ denotes feature concatenation. This
representation characterizes both the semantic persistence of the
original phrase and clinically meaningful changes induced by localized
visual intervention.

\subsubsection{Counterfactual Alignment Cues}
\label{sec:alignment_cues}

Counterfactual alignment cues characterize how image--phrase compatibility changes after localized disruption of the candidate region. Let $\Phi(I,c_i)$ denote an image--phrase compatibility function that
quantifies how well the visual input aligns with the semantic content
of phrase $c_i$. The alignment scores before and after intervention are
defined as
\begin{equation}
a_i^{0}
=
\Phi(I,c_i),
\qquad
a_i'
=
\Phi
\left(
I_i',c_i
\right).
\label{eq:alignment_scores}
\end{equation}

The intervention-induced alignment variation is defined as
\begin{equation}
\Delta a_i
=
a_i^{0}
-
a_i',
\label{eq:alignment_variation}
\end{equation}
and the corresponding relative variation is
\begin{equation}
\Delta a_i^{\mathrm{r}}
=
\frac{
a_i^{0}-a_i'
}{
\left|a_i^{0}\right|+\epsilon
},
\label{eq:relative_alignment_variation}
\end{equation}
where $\epsilon$ is a small constant introduced for numerical
stability.

In addition to the original alignment strength and
intervention-induced variation, the alignment representation includes
a grounding descriptor $g_i$ characterizing the localization output and
a region-availability indicator $r_i$ specifying whether a valid localized region is identified. The counterfactual alignment
cue is therefore represented as
\begin{equation}
\mathbf z_i^{\mathrm{a}}
=
\left[
a_i^{0};
a_i';
\Delta a_i;
\Delta a_i^{\mathrm{r}};
g_i;
r_i
\right].
\label{eq:alignment_cues}
\end{equation}

Together, these observations characterize both the initial
image--phrase correspondence and its sensitivity to localized
counterfactual intervention. Importantly, alignment variation is treated as an auditing cue rather than as a deterministic criterion for visual relevance.

\subsection{Cue Normalization and Hallucination Risk Modeling}
\label{sec:risk_modeling}

The counterfactual auditing representation $\mathbf z_i$ comprises
heterogeneous response and alignment cues whose numerical scales may
vary across different VLMs and datasets. Directly combining these raw cue values may therefore introduce scale biases into VIH risk estimation. To improve comparability, CLEAR augments each cue
dimension with a percentile-normalized counterpart.

For each VLM--dataset setting, let $\mathcal U$ denote an unlabeled
reference set of counterfactual auditing representations. For the
$j$-th cue dimension, the percentile-normalized value is defined as
\begin{equation}
\widetilde z_{i,j}
=
\frac{1}{|\mathcal U|}
\sum_{\mathbf z_u\in\mathcal U}
\mathbb{I}
\left(
z_{u,j}
\leq
z_{i,j}
\right),
\label{eq:percentile_normalization}
\end{equation}
where $\mathbb{I}(\cdot)$ denotes the indicator function. Applying Eq.~\eqref{eq:percentile_normalization} to each dimension of
$\mathbf z_i$ yields the normalized cue vector
$\widetilde{\mathbf z}_i$. The final representation is obtained by
concatenating the original and normalized cues:
\begin{equation}
\mathbf x_i
=
\operatorname{Concat}
\left[
\mathbf z_i;
\widetilde{\mathbf z}_i
\right].
\label{eq:normalized_representation}
\end{equation}
This normalization depends only on the empirical distribution of the
auditing cues and does not require VIH labels or ground-truth reports. It reduces scale discrepancies across different
VLMs and datasets while retaining both the original cue values and
their relative positions within the reference distribution.

Based on $\mathbf x_i$, CLEAR estimates the phrase-level VIH risk
through a lightweight risk estimator $M$:
\begin{equation}
\widehat h_i
=
M
\left(
\mathbf x_i
\right),
\label{eq:hallucination_risk}
\end{equation}
where $\widehat h_i\in[0,1]$ denotes the estimated risk that phrase
$c_i$ is visually irrelevant. CLEAR therefore estimates VIH risk
from the joint response and alignment cues and their normalized
counterparts, with $\widehat h_i$ interpreted as a phrase-level VIH
risk score.

\section{Experiments}

\subsection{Public Evaluation Datasets}
To evaluate phrase-level visually irrelevant hallucination detection across diverse clinical scenarios, we conduct extensive experiments on four representative public datasets spanning two medical vision--language tasks: medical visual question answering (Med-VQA) and radiology report generation (RRG). 

For the RRG task, we use MIMIC-CXR~\cite{johnson2019mimic} and IU-Xray~\cite{demner2015preparing}. MIMIC-CXR is a large-scale chest radiography dataset containing chest X-ray images paired with semi-structured radiology reports. We follow the official patient-level data splits and use only the test split for evaluation. IU-Xray contains chest radiographs paired with 2,955 radiology reports, and we similarly use only its predefined test split. These datasets enable evaluation of visually irrelevant hallucinations in free-form radiology report generation, where a single generated response may contain multiple clinical phrases with different degrees of visual support. 

For the Med-VQA task, we use VQA-RAD~\cite{lau2018dataset} and MIMIC-Diff-VQA~\cite{bae2024mimic}. VQA-RAD contains 2,244 question--answer pairs associated with 314 radiology images. Since our study focuses on explicitly generated and visually assessable clinical phrases, we evaluate hallucination detection on the 200 open-ended questions in its test split. MIMIC-Diff-VQA is a large-scale benchmark covering clinically relevant question categories such as abnormality, location, and presence. Following the same evaluation principle, we exclude categories with fixed-template answers, such as \textit{view} questions, as well as binary \textit{yes/no} questions. This filtering results in 5,610 open-ended test samples. For all four datasets, the evaluated VLM first generates a free-form response from the corresponding image and query or reporting instruction. The generated response is subsequently decomposed into atomic, visually assessable clinical phrases as described in Definition~\ref{def:VIH}. Hallucination detection is then performed at the phrase level by determining the risk that each generated phrase is insufficiently supported by, or inconsistent with, the visual input.

\begin{table*}[!t]
\centering
\caption{HALLUCINATION DETECTION PERFORMANCE  OF CLEAR AND EIGHT COMPETING METHODS ON OPEN-ENDED VQA AND RRG  BENCHMARKS. BEST RESULTS ARE SHOWN IN \textbf{BOLD}, AND THE SECOND-BESTRE SULTS ARE \underline{UNDERLINED}.Statistical significance using
pairedt-tests are reported,wherenumbers in \textbf{\textcolor{blue}{blue}} and \textcolor{red}{red}
indicate p-values $<$ 0.05 and $<$ 0.01, respectively.AUC and PRC denote Area Under the Receiver Operating Characteristic Curve (AUROC) and Area Under the Precision-Recall Curve (AUPRC), respectively.}
\label{tab:hallucination-single-column}
\small 
\setlength{\tabcolsep}{4.5pt} 
\begin{tabular}{l ccc ccc ccc ccc}
\toprule
\multirow{2}{*}{\textbf{Method}} & \multicolumn{3}{c}{{MIMIC-CXR}} & \multicolumn{3}{c}{{IU-Xray}}  & \multicolumn{3}{c}{{VQA-RAD}}  & \multicolumn{3}{c}{{MIMIC-Diff-VQA}} \\
\cmidrule(lr){2-4} \cmidrule(lr){5-7} \cmidrule(lr){8-10} \cmidrule(lr){11-13}
& {AUC} $\uparrow$ & {PRC} $\uparrow$ & {F1} $\uparrow$ & {AUC} $\uparrow$ & {PRC} $\uparrow$ & {F1} $\uparrow$ & {AUC} $\uparrow$ & {PRC} $\uparrow$ & {F1} $\uparrow$& {AUC} $\uparrow$ & {PRC} $\uparrow$ & {F1} $\uparrow$\\
\midrule

\rowcolor{gray!10} \multicolumn{13}{c}{{ MedGemma-4B-it~\cite{sellergren2025medgemma}}} \\
AvgProb &67.33 &76.00 & 66.88& 55.86&71.70 &55.57 &60.06 &62.80 &56.12 &47.39 & 78.01&69.60\\
AvgEnt &68.06 &78.84 &68.90 & 60.84&74.23 &59.50 &65.69 &78.12 & 64.97&54.94 &80.95 &75.71\\
MaxProb &65.28 & 75.36&66.97 &58.74 &73.78 &57.98 &61.58 &60.12 &54.92 &49.60 &79.20 &71.52\\
MaxEnt &66.21 &77.62 &68.33 & 61.59& 75.31&64.20 &65.06 & 78.21& 72.68& 60.87&80.71 &78.29\\
Cross-Checking   & 69.88& 78.00& 68.89& 61.39& 74.17&68.62 &70.80 & 81.35&75.83 &62.81 &84.18 &80.96\\
RadFlag & 70.02& 80.19& 72.03& 65.34& 75.31&69.39 &69.33 &80.82 &74.59 &62.29 &84.69 &81.48\\
SE &73.36 &82.24 &75.14&72.69 & 79.25& 72.51  &70.58 &80.57 &75.62 &65.99 &88.41 &84.60\\
{VASE} &\underline{76.60} & \underline{85.10}& \underline{80.93}&\underline{75.80} &\underline{80.32} &\underline{76.10} &\underline{74.62} &\underline{87.02} &\underline{78.07} &\underline{67.28} &\underline{89.16} &\underline{85.74}\\

Ours (MedGemma) & \textbf{\textcolor{blue}{79.75}} & \textbf{\textcolor{blue}{90.23}} & \textbf{\textcolor{blue}{84.07}}& \textbf{\textcolor{blue}{76.00}}& \textbf{\textcolor{blue}{81.40}}& \textbf{\textcolor{blue}{79.79}}  & \textbf{\textcolor{blue}{77.46}} & \textbf{\textcolor{blue}{89.13}} & \textbf{\textcolor{blue}{78.47}} & \textbf{\textcolor{blue}{69.48}} & \textbf{\textcolor{blue}{92.86}} & \textbf{\textcolor{blue}{89.56}} \\

\midrule

\rowcolor{gray!10} \multicolumn{13}{c}{{LLava-Med~\cite{li2023llava}}} \\
AvgProb &59.24 &74.17 & 69.50 &50.95 &74.32 &66.77 &54.42 &73.95 &71.05 & 49.89 &79.56 &75.78\\
AvgEnt &61.63 &78.18& 71.78&46.01 &73.54 &66.13 & 61.58&77.95 & 74.89& 53.90 &82.09 &76.96\\
MaxProb &61.73 &78.59 &71.12 &52.56 &76.01 &71.53 &50.45 &68.73 &69.76 &46.98 & 80.58&74.72\\
MaxEnt &64.46 &79.38 &79.53 &49.81 &67.64 & 68.79&63.33 &73.81 &75.09 &54.72 &83.85 &77.12\\
Cross-Checking   &67.49 &84.78 &81.58 &62.52 &78.16 &74.25 & 70.50& 77.09&76.25 & 56.65&85.63 &77.29\\
RadFlag &70.33 &86.97 &83.42 &62.68 &78.12 &75.58 & 73.22&80.07 &78.24 & 59.29& 88.50&80.40\\
SE &72.85 &88.41 &87.59 &65.01 &80.31 &76.42 & 74.03&81.13 &79.87 & 63.72&\underline{90.58} &81.93\\
VASE &\underline{74.30} &\underline{90.45} &\underline{89.35} & \underline{70.02}&\underline{84.24} & \underline{77.34}&\textbf{75.55} &\textbf{82.79} &\textbf{83.78} &\underline{67.63} & {89.84}&\underline{83.35}\\
Ours (LLava-Med) & \textbf{\textcolor{blue}{79.12 }}& \textbf{\textcolor{blue}{94.83}} & \textbf{\textcolor{blue}{91.22}}& \textbf{75.67} & \textbf{\textcolor{blue}{89.29}} & \textbf{\textcolor{blue}{88.83}} & \underline{\textcolor{red}{73.30}} & \underline{\textcolor{blue}{81.27}} & \underline{\textcolor{blue}{82.99}} & \textbf{\textcolor{blue}{73.94}} & \textbf{90.92} & \textbf{\textcolor{red}{85.90}} \\


\midrule


\rowcolor{gray!10} \multicolumn{13}{c}{{ Lingshu-7B~\cite{xu2025lingshu}}} \\
AvgProb &40.68 & 71.54&45.75&44.56 &70.10 & 50.87 &53.59 &73.08 &65.36 & 50.18&78.37 &74.64\\
MaxProb &54.46 & 75.72&50.47 & 50.00& 77.39& 59.90 & 56.53&79.51 &68.40  &56.68 &80.18 & 77.86\\
AvgEnt &42.33 & 72.37&47.57& 48.82&74.32 & 58.94 & 54.58&76.60 &67.95 &54.03 &79.82 &76.49\\
MaxEnt &55.28 &75.29 &51.27 &51.18 &78.12 & 56.33 &58.90 &80.20 &72.34 &60.96 &81.04 &78.10\\
Cross-Checking&59.58 &80.31 &54.82 & 60.65&79.72 &60.17  & 61.15&84.91 &73.27 & 61.76&81.98 &78.09\\
RadFlag  &61.18 &85.32 & 57.55&67.65 &85.61 & 63.68 &67.34 &89.51 &75.65 &61.01 &82.46 &79.28\\
SE &62.84 & \textbf{89.81} &56.35 &70.47 &86.40 & 65.25 &71.35 & \underline{91.07}&76.77 &62.76 &83.18 &80.04\\
{VASE} & \underline{65.22} &\underline{88.16} & \underline{60.02} &\underline{74.14} &\textbf{88.22} &\underline{71.82} &\textbf{72.19} &\textbf{92.21} &\textbf{77.34} &\underline{63.15}  &\underline{84.94} &\underline{80.18}\\
Ours (Lingshu-7B) &  \textbf{74.96} & \textcolor{blue}{87.93} & \textbf{\textcolor{blue}{78.49}} & \textbf{\textcolor{blue}{77.59 }}& \textbf{\textcolor{blue}{90.84}} & \textbf{\textcolor{blue}{74.78}} & \underline{\textcolor{red}{72.08}} & \textcolor{blue}{90.39} & \underline{77.04} & \textbf{\textcolor{blue}{68.91}} & \textbf{\textcolor{blue}{89.98}} & \textbf{85.33} \\

\midrule

\rowcolor{gray!10} \multicolumn{13}{c}{{Qwen-7B~\cite{bai2023qwen}}} \\
AvgProb &52.46 &73.28 &55.12 & 55.04& 68.68&55.19 & 60.22& 74.22& 70.69&48.31 &76.28 &69.77\\
MaxProb &54.28 &75.15 &58.64 &57.21 &71.89 &55.66 &63.69 &80.92 &73.08 &54.52 &79.03 &74.89\\

AvgEnt &54.69 &75.80 &58.17 & 52.36&66.73 & 51.82 &55.35 & 67.98&69.11 &50.96 &78.64 &73.50\\
MaxEnt &57.32 &78.51 &59.63  & 56.77&70.62 & 57.38 &60.75 &76.64 &70.72 &58.64 &82.58 &77.45\\
Cross-Checking   &56.43 &77.35 &59.26 & 57.83&71.73 & 56.74 & 65.12& 83.66& 73.52&61.31 &83.92 &81.02\\
RadFlag &58.57 &78.71 &60.30 &61.30 &76.42 &59.40 & 66.44&85.19 &75.27 &62.42 &84.96 &83.47\\
SE &60.96 &80.88 &62.43 &66.91 & 80.91&60.35 &72.52 &89.51 &76.73 &64.18 &86.17 &85.51\\
VASE &\underline{62.50} &\underline{81.99} &\underline{65.15} &\underline{68.08} &\underline{81.93} & \underline{60.64} &\underline{76.88} &\underline{94.15} &\underline{78.05} &\underline{64.83} &\underline{86.24} &\underline{85.93}\\

Ours (Qwen-7B) & \textbf{\textcolor{blue}{68.40}} & \textbf{\textcolor{blue}{85.43}} & \textbf{\textcolor{blue}{70.74}} & \textbf{\textcolor{blue}{69.01}} & \textbf{\textcolor{blue}{82.65}} & \textbf{\textcolor{blue}{62.75}} & \textbf{\textcolor{red}{76.91}} & \textbf{\textcolor{red}{95.16}} & \textbf{\textcolor{red}{78.73}} & \textbf{\textcolor{blue}{68.40}} & \textbf{\textcolor{blue}{89.93}} & \textbf{\textcolor{blue}{89.38}} \\

\bottomrule
\end{tabular}
\end{table*}

\subsection{GT-Derived Controlled VIH Benchmark}
Evaluation on public datasets reflects naturally generated clinical responses, but the resulting hallucination labels may depend on heterogeneous annotation procedures, reference representations, and dataset-specific matching rules. Moreover, existing medical vision--language datasets do not provide a universally accepted gold standard specifically for phrase-level visually irrelevant hallucinations. To provide a complementary evaluation with transparent and reproducible visual-support labels, we construct a \emph{GT-derived controlled VIH benchmark}. 

The benchmark contains 200 atomic and visually assessable clinical phrases, comprising 100 visually supported control phrases and 100 paired visually irrelevant phrases. Each control phrase is derived from a ground-truth or reference-supported clinical statement. Its paired visually irrelevant counterpart is generated through a minimal, clinically constrained semantic edit that modifies one clinically meaningful property while preserving the remaining linguistic structure as much as possible. The controlled edits include laterality alteration, negation inversion, severity alteration, and clinical entity substitution.

Importantly, an edited phrase is retained only when the modification changes its visual-support status under Definition~\ref{def:VIH}. Specifically, the resulting phrase must remain clinically plausible and visually assessable while being insufficiently supported by, or inconsistent with, the corresponding image. The resulting pairs therefore preserve closely matched linguistic form while differing in whether their asserted clinical content is supported by the visual input. 

This controlled construction complements evaluation on naturally generated responses by reducing the influence of dataset-specific annotation protocols and superficial linguistic differences. It therefore provides a direct test of whether the proposed method can distinguish visually supported from visually irrelevant clinical phrases under a transparent and reproducible labeling setting. The complete construction and validation procedures are provided in the supplementary material.

To ensure label validity, all 200 image--phrase pairs and their assigned visual-support labels were subsequently reviewed by two clinicians with expertise in medical image interpretation.

\subsection{Evaluation Metrics}
To quantitatively assess the performance of the proposed hallucination detection framework, we adopt three comple-
mentary metrics that evaluate the alignment between estimated scores and ground-truth clinical labels.

We employ three complementary metrics following~\cite{farquhar2024detecting,liao2025vision}.The primary metric is the Area Under the Receiver Operating Characteristic curve (AUROC), which measures the probability that a randomly selected hallucinated phrase receives a higher hallucination score than a randomly selected factual phrase. AUROC provides a threshold-independent as sessment of the framework’s ranking ability across the entire sensitivity-specificity spectrum. Complementary to this, we employ the Area Under the Precision-Recall Curve (AUPRC), which is particularly effective for medical datasets where the distribution of factual assertions often significantly outweighs that of hallucinated content. A higher AUPRC indicates that the framework can maintain high precision even at high recall levels, thereby minimizing the risk of false clinical alarms. Finally, we report the F1-score to evaluate the classification
performance. The F1-score represents the harmonic mean of precision and recall, providing a balanced measure of the
framework’s ability to accurately isolate hallucinations while preserving clinical facts. For each metric, a higher value signifies superior detection efficacy. By integrating these three metrics, we ensure a comprehensive evaluation that accounts for both the ranking quality and the practical classification reliability of the hallucination detection procedure.

\subsection{Implementation Details}
CLEAR was implemented using PyTorch and scikit-learn, and
all experiments were conducted on two NVIDIA RTX L20 GPUs.
For each phrase \(c_i\), MAIRA-2 was used as the
phrase-conditioned grounding model to propose candidate support
regions \(\mathcal B_i\). These regions were treated only as
auditing targets rather than verified evidence of factual support.
When no valid localized region was returned, the full image was
used as the fallback intervention region. Unless otherwise
specified, regional Gaussian blur with radius 20 was applied in
the main experiments, with the same setting used across all
evaluated VLMs and datasets. To ensure the reliability of the similarity metrics, we utilized a pre-trained CLIP-based medical vision. We generated sentence-level hallucination labels using two independent large language models by comparing each generated response against the corresponding reference answer. Only labels on which both models reached consensus were retained as the final hallucination annotations.

\subsection{Comparison with State-of-the-Art Methods}
\label{sec:sota_comparison}

Table~\ref{tab:hallucination-single-column} compares CLEAR with representative uncertainty-, consistency-, and verification-based hallucination detectors on four benchmarks covering RRG and Med-VQA. The evaluation includes both a general-purpose VLM and three medical VLMs, allowing us to examine whether the proposed evidence auditing strategy generalizes across different backbone types. Overall, CLEAR achieves the best AUROC in most backbone--dataset combinations and provides consistent improvements on both long-form and short-form clinical generation tasks.

\subsubsection{Performance on the general-purpose VLM.}
We first evaluate CLEAR using Qwen-VL, which is not specifically pre-trained for medical image understanding. CLEAR substantially improves hallucination detection on MIMIC-CXR, increasing AUROC from $62.50\%$ with VASE to $68.40\%$, AUPRC from $81.99\%$ to $85.43\%$, and F1 from $65.15\%$ to $70.74\%$. It also achieves the
best performance on MIMIC-Diff-VQA, with an AUROC of $68.40\%$, an AUPRC of $89.93\%$, and an F1-score of $89.38\%$. On VQA-RAD, CLEAR remains competitive with the strongest baseline while obtaining the highest AUROC and AUPRC. These results indicate that the proposed counterfactual auditing cues remain informative even when the underlying VLM lacks domain-specific medical pre-training.

\subsubsection{Performance on medical VLMs.}
CLEAR further demonstrates strong performance across MedGemma, LLaVA-Med, and Lingshu, which differ in architecture, pre-training data, and medical capability. The improvements are particularly consistent on the two RRG benchmarks. On MIMIC-CXR with MedGemma, CLEAR improves AUROC from $76.60\%$ to $79.75\%$, AUPRC from
$85.10\%$ to $90.23\%$, and F1 from $80.93\%$ to $84.07\%$ relative to VASE. With LLaVA-Med, it increases AUROC from $74.30\%$ to $79.12\%$ and AUPRC from $90.45\%$ to $94.83\%$. The gains are even larger for Lingshu in terms of AUROC and F1, where CLEAR improves MIMIC-CXR AUROC from $65.22\%$ to $74.96\%$ and F1 from $60.02\%$
to $78.49\%$. CLEAR also achieves the highest AUROC on IU-Xray for all three medical backbones. The consistent improvements across different medical VLMs suggest that localized counterfactual auditing
does not depend on a particular model architecture or level of medical specialization.

The results on VQA are more varied. CLEAR achieves the best overall performance on MIMIC-Diff-VQA across all evaluated backbones and obtains the strongest results on VQA-RAD with MedGemma and Qwen-VL.
For LLaVA-Med and Lingshu on VQA-RAD, CLEAR remains competitive but does not dominate every metric. This difference is expected because open-ended VQA answers are generally shorter than radiology reports,
providing fewer textual phrases and less contextual information for counterfactual response analysis. Nevertheless, the strong results on MIMIC-Diff-VQA show that CLEAR remains effective for short-form
generation when the answer contains clinically meaningful descriptive content.

\subsubsection{Comparison across hallucination detector families.}
A clear performance progression can be observed across different hallucination detection paradigms. Token-level probability and entropy baselines, including AvgProb, MaxProb, AvgEnt, and MaxEnt, generally
provide the weakest results because they assess generation confidence without directly measuring visual support. Cross-Checking and RadFlag improve performance by introducing response-level or external
verification, while Semantic Entropy and VASE further benefit from sampling-based semantic consistency and multimodal uncertainty. However, these approaches still summarize hallucination risk primarily
through one dominant reliability signal.

In contrast, CLEAR jointly models counterfactual response
and alignment cues under localized intervention. Response cues
characterize semantic persistence and clinically meaningful
changes in the regenerated response, whereas alignment cues
characterize the original image--phrase compatibility and its
variation after intervention. The results support the central
hypothesis of this work: phrase-level VIH risk is more reliably
characterized through complementary response and alignment
observations than through generation uncertainty, response
consistency, or visual grounding alone.
\begin{table}[t]
\centering
\caption{
Hallucination detection performance on the GT-derived controlled evaluation
set. The best result in
each column is shown in \textbf{bold}.
}
\label{tab:controlled_detection_results}
\scriptsize
\setlength{\tabcolsep}{4pt}
\begin{tabular}{l  ccc ccc}
\toprule
\multirow{2}{*}{\textbf{Method}}
& \multicolumn{3}{c}{\textbf{RRG}}
& \multicolumn{3}{c}{\textbf{VQA}} \\
\cmidrule(lr){2-4}
\cmidrule(lr){5-7}
& \textbf{AUROC} & \textbf{AUPRC} & \textbf{F1}
& \textbf{AUROC} & \textbf{AUPRC} & \textbf{F1} \\
\midrule

RadFlag
& 76.69 & 76.51 & 79.83
& 79.03 & 85.82 & 85.88 \\

Cross-Checking
& 74.62 & 79.66 & {80.62}
& 73.67 & 82.18 & 79.49 \\

\textbf{CLEAR}

& \textbf{80.10}
& \textbf{89.08}
& \textbf{85.28}
& \textbf{82.11}
& \textbf{87.43}
& \textbf{91.55} \\

\bottomrule
\end{tabular}
\end{table}

\begin{figure*}
    \centering
    \includegraphics[width=1\linewidth]{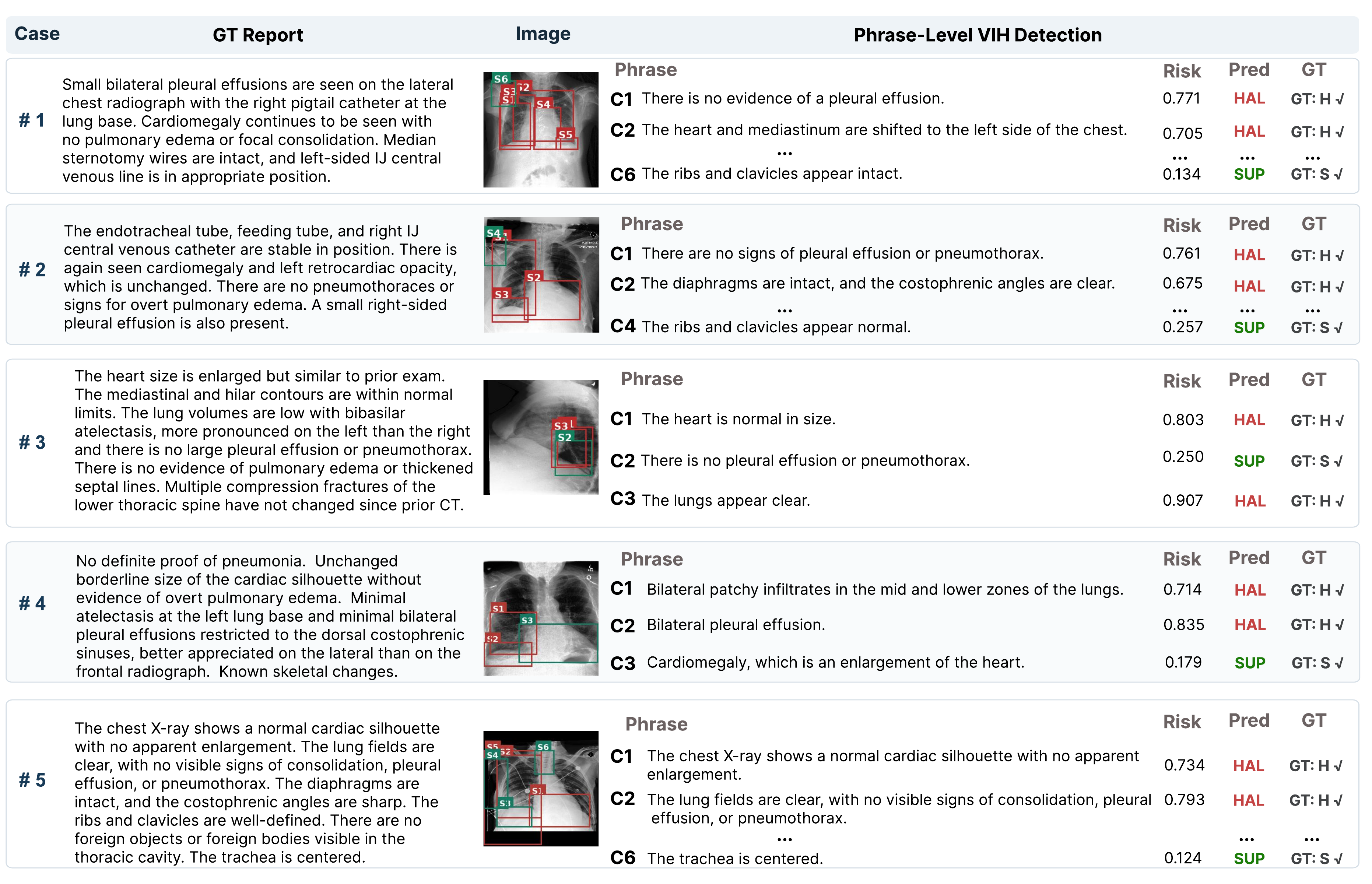}
    \caption{
Qualitative phrase-level hallucination detection examples of CLEAR. 
For each radiology image, multiple generated clinical phrases are grounded to their corresponding visual regions and independently assigned hallucination risk scores. 
Red boxes indicate phrases predicted as hallucinated, while green boxes indicate phrases predicted as supported. 
}
    \label{fig:placeholder}
\end{figure*}

\subsection{Evaluation on the VIH Benchmark}
\label{sec:controlled_results}

Table~\ref{tab:controlled_detection_results} reports the evaluation on the proposed GT-derived controlled set. Compared with existing medical hallucination benchmarks, this evaluation adopts a transparent label construction protocol in which hallucinated statements are generated through controlled semantic edits of GT-supported clinical statements. CLEAR achieves the best AUROC and AUPRC on both the RRG and VQA subsets, improving AUROC from 76.69\% to 80.10\% on RRG and from 79.03\% to 82.11\% on VQA relative to the strongest competing baseline. The substantial improvement in AUPRC further indicates that CLEAR more effectively distinguishes GT-supported statements from their clinically inconsistent counterparts under a balanced evaluation setting.

Notably, the controlled evaluation is considerably more challenging than conventional benchmark evaluation because each hallucinated statement differs from its corresponding GT-supported control by only a minimal clinically meaningful edit. The consistent performance improvement therefore suggests that CLEAR captures clinically relevant evidence inconsistency rather than relying on superficial lexical differences or a particular hallucination annotation protocol. Together with the results on public benchmarks, these findings demonstrate that the proposed evidence auditing framework generalizes well across both naturally generated and controlled hallucination scenarios.

\subsection{Ablation Study} Table~\ref{tab:clear_evidence_ablation} examines the contributions of the counterfactual
response cues, alignment cues, and percentile normalization.
Response cues alone achieve an average AUROC of 70.87 on RRG
and 68.02 on Med-VQA, whereas alignment cues alone achieve
65.46 and 65.18, respectively. This result indicates that
semantic persistence and clinically meaningful response changes
after localized intervention provide informative signals of
evidence-insensitive generation. Nevertheless, combining the
response and alignment cues consistently improves performance,
demonstrating that the two cue families capture complementary
aspects of phrase behavior.

The ablation results also support the use of multi-evidence modeling rather than a deterministic counterfactual rule. Although textual persistence is individually effective, it does not reliably distinguish hallucinations from valid but stable statements, such as global findings, negated observations, and common reporting templates. The additional evidence streams provide the contextual information required to interpret such ambiguous counterfactual behavior.

\begin{table}[ht]
\centering
\caption{
Ablation study of CLEAR. Resp., Align., and Norm. denote the counterfactual response cues, counterfactual alignment cues, and percentile normalization, respectively. Results are macro-averaged across the evaluated VLM--dataset settings within the RRG and Med-VQA tasks. The best result in each column is shown in \textbf{bold}.
}
\label{tab:clear_evidence_ablation}
\scriptsize
\setlength{\tabcolsep}{2.7pt}
\begin{tabular}{ccc ccc ccc}
\toprule
\multicolumn{3}{c}{\textbf{CLEAR Components}}
& \multicolumn{3}{c}{\textbf{RRG}}
& \multicolumn{3}{c}{\textbf{VQA}}
\\
\cmidrule(lr){1-3}
\cmidrule(lr){4-9}
\textbf{Resp.}
& \textbf{Align.}
& \textbf{Norm.}
& \textbf{AUROC}
& \textbf{AUPRC}
& \textbf{F1}
& \textbf{AUROC}
& \textbf{AUPRC}
& \textbf{F1}
\\
\midrule

\cmark & \xmark  & \xmark
& 70.87 & 83.49 & 72.48
& 68.02 & 85.01 & 76.18
\\

\xmark & \cmark & \xmark
& 65.46 & 78.71 & 66.41
& 65.18 & 80.79 & 67.80
\\

\cmark & \cmark  & \xmark
& 72.02 & 84.13 & 74.44
& 70.89 & 86.60 & 80.84
\\

\cmark & \cmark & \cmark
& \textbf{75.06}
& \textbf{87.83}
& \textbf{78.84}
& \textbf{72.56}
& \textbf{89.95}
& \textbf{83.43}
\\

\bottomrule
\end{tabular}
\end{table}

\begin{figure}[t]
    \centering
    \includegraphics[width=\linewidth]{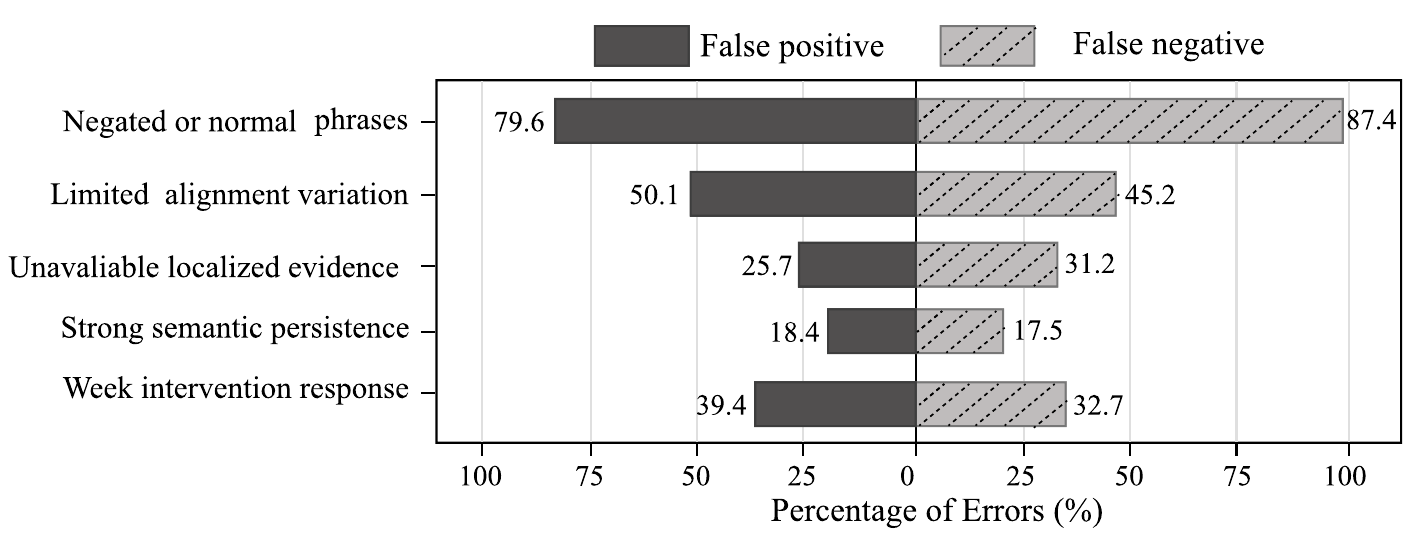}
    \caption{
    Hard-case analysis of CLEAR. Bars report the prevalence of five
    non-exclusive phrase and evidence patterns among false-positive and
    false-negative cases. Negated or normal phrases are the most prevalent
    pattern, followed by limited counterfactual alignment variation.
    }
    \label{fig:hard_case_patterns}
\end{figure}

\subsection{Hard-Case and Auditing Cue Analysis}
\label{sec:error_attribution}

To better characterize the remaining errors of CLEAR, we examine both the prevalence of challenging phrase and evidence patterns and the relative
contributions of the two counterfactual auditing-cue families.

As shown in Fig.~\ref{fig:hard_case_patterns}, the analyzed patterns are non-exclusive and may co-occur within the same error case. Negated or normal
phrases constitute the most prevalent pattern, appearing in 79.6\% of false positives and 87.4\% of false negatives. Limited counterfactual alignment
variation is the second most frequent pattern, occurring in 50.1\% and 45.2\% of the two error types, respectively. Weak intervention responses are also
common, particularly among false positives (39.4\% versus 32.7\%), while localized evidence is unavailable in 25.7\% of false positives and 31.2\% of
false negatives. These results indicate that templated negative phrases and findings producing weak or insufficiently localized counterfactual changes
remain particularly challenging. Strong semantic persistence is the least frequent pattern, occurring in 18.4\% of false positives and 17.5\% of false
negatives; nevertheless, its presence among both error types confirms that persistence alone cannot be treated as definitive evidence of visual support.

\begin{figure}[t]
    \centering
    \includegraphics[width=0.95\linewidth]{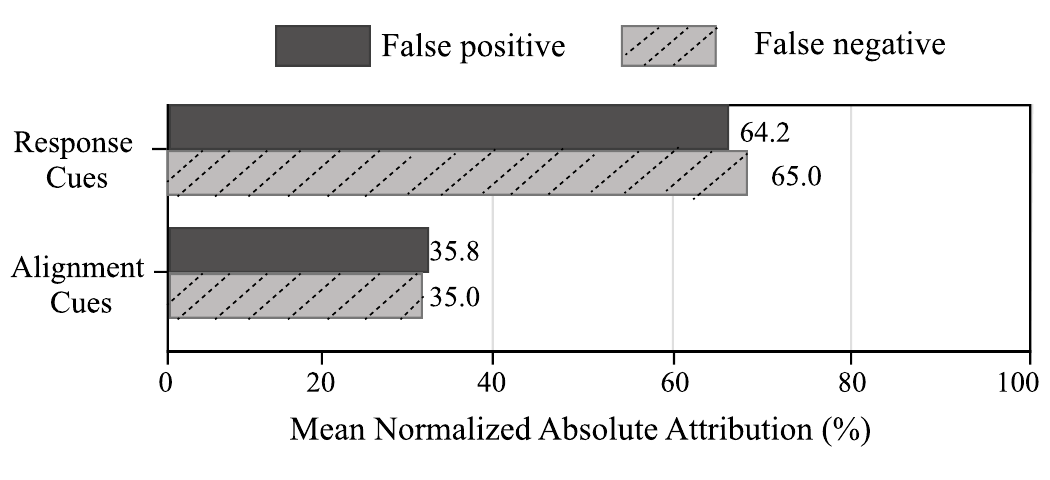}
    \caption{
    Counterfactual auditing-cue attribution of CLEAR. Absolute feature attributions are aggregated into response cues, including semantic persistence and clinically meaningful response changes, and alignment cues, capturing image--phrase compatibility and its intervention-induced variation. Bars show the mean normalized absolute attribution for false-positive and false-negative cases. Both error types exhibit similar attribution patterns, with response cues contributing approximately 65\% and alignment cues approximately 35\%.
    }
    \label{fig:evidence_attribution}
\end{figure}

Fig.~\ref{fig:evidence_attribution} further reports the mean normalized absolute attribution of the two counterfactual auditing-cue families.
Response cues account for 64.2\% of the attribution among false positives and 65.0\% among false negatives, whereas alignment cues account for 35.8\% and
35.0\%, respectively. The nearly identical distributions indicate that the two error types cannot be distinguished by a simple shift in relative reliance
between response- and alignment-based evidence. Instead, the learned estimator primarily relies on counterfactual response behavior in both cases, while
image--phrase alignment provides a substantial complementary contribution. These findings support the joint modeling of the two cue families, while also
suggesting that further improvements require more discriminative cues within each family, particularly for negated or normal phrases and interventions that
induce only weak observable changes.

\section{Discussion}
\subsection{Fine-Grained Auditing of Visual Relevance}

The results support phrase-level visual-relevance auditing as a complementary approach to conventional uncertainty-, consistency-, and verifier-based hallucination detection. CLEAR achieves consistent improvements across RRG and Med-VQA, while the controlled VIH benchmark further shows that the framework can distinguish visually relevant and irrelevant clinical statements even when they differ only by subtle clinically meaningful edits. These findings suggest that response-level reliability alone is insufficient for identifying which individual phrases within an otherwise plausible response are problematic.

The ablation and hard-case analyses further clarify the role of the proposed auditing cues. Counterfactual response cues are individually more informative, but combining them with image--phrase alignment consistently improves performance, indicating complementary information across the two cue families. At the same time, negated or normal phrases and cases with weak intervention-induced changes remain challenging, reinforcing that neither grounding, semantic persistence, nor alignment variation should be interpreted as a deterministic criterion for VIH.

\subsection{Robustness to Counterfactual Intervention}

To examine whether CLEAR depends on the specific form or strength of the localized intervention, we evaluate both alternative perturbation operators and different perturbation scales. Specifically, we compare Gaussian blur, binary masking, mean-value replacement, and noise replacement, and further vary the Gaussian blur radius from 5 to 20. As shown in Tables~\ref{tab:perturbation_robustness} and~\ref{tab:blur_radius_mrg}, CLEAR maintains highly comparable VIH detection performance across different perturbation settings and backbone models.

No single perturbation operator consistently achieves the best performance across all datasets and VLM backbones. Gaussian blur, masking, mean replacement, and noise replacement yield only minor performance differences. A similar trend is observed when varying the blur radius: although individual metrics occasionally favor different perturbation scales, no consistent preference emerges across datasets or backbones. These results indicate that CLEAR is not strongly dependent on either a particular perturbation operator or precise perturbation intensity.

This robustness is consistent with the role of localized counterfactual intervention in CLEAR. The intervention is used primarily as a probe rather than as a deterministic criterion for VIH. CLEAR instead models the resulting changes in generated responses and image--phrase alignment through complementary auditing cues. As long as the candidate region is meaningfully disrupted while the surrounding image context is largely preserved, different perturbation operators and scales can produce sufficiently informative counterfactual patterns for stable phrase-level VIH estimation. This relative insensitivity also reduces the need for extensive perturbation-specific tuning across datasets and VLM architectures.

\begin{table}[t]
\centering
\caption{
Robustness analysis of CLEAR under different localized counterfactual
perturbations across backbone models. Blur, Mask, Mean, and Noise denote
alternative operators for constructing counterfactual images. Comparable
performance across these perturbations indicates that CLEAR is not strongly
dependent on a specific intervention strategy.
}
\label{tab:perturbation_robustness}
\scriptsize
\setlength{\tabcolsep}{3.2pt}
\begin{tabular}{ll ccc ccc}
\toprule
\multirow{2}{*}{\textbf{Backbone}}
& \multirow{2}{*}{\textbf{Perturbation}}
& \multicolumn{3}{c}{\textbf{MIMIC-CXR}}
& \multicolumn{3}{c}{\textbf{IU-Xray}} \\
\cmidrule(lr){3-5}
\cmidrule(lr){6-8}
&
& \textbf{AUROC} & \textbf{AUPRC} & \textbf{F1}
& \textbf{AUROC} & \textbf{AUPRC} & \textbf{F1} \\
\midrule

\multirow{4}{*}{MedGemma}
& Blur
& \textbf{79.75}
& \textbf{90.23}
& 84.07
& 76.00
& 81.40
& 79.79 \\

& Mask
& 79.60
& 90.09
& 84.06
& 76.52
& 81.72
& \textbf{80.01} \\

& Mean
& 79.60
& 90.08
& \textbf{84.12}
& 76.55
& \textbf{82.01}
& 79.95 \\

& Noise
& 79.62
& 90.11
& 84.11
& \textbf{76.59}
& 81.83
& 79.89 \\

\midrule

\multirow{4}{*}{LLaVA-Med}
& Blur
& 79.12
& 94.83
& \textbf{91.22}
& 75.67
& 89.29
& 88.83 \\

& Mask
& 79.27
& 94.76
& 91.07
& 77.94
& 89.46
& \textbf{88.90} \\

& Mean
& 79.33
& 94.86
& 91.02
& \textbf{78.40}
& \textbf{89.53}
& \textbf{88.90} \\

& Noise
& \textbf{79.57}
& \textbf{94.99}
& 91.14
& 76.64
& 89.32
& 88.83 \\

\bottomrule
\end{tabular}
\end{table}
\begin{table}[t]
\centering
\caption{
Sensitivity analysis of CLEAR with respect to the Gaussian blur radius
on the MIMIC-CXR and IU-Xray benchmarks. The results show that CLEAR
maintains stable detection performance across different blur scales.
The best result for each backbone and evaluation metric is highlighted
in \textbf{bold}.
}
\label{tab:blur_radius_mrg}
\scriptsize
\setlength{\tabcolsep}{3.2pt}

\begin{tabular}{ll ccc ccc}
\toprule
\multirow{2}{*}{\textbf{Backbone}}
& \multirow{2}{*}{\textbf{Blur Radius}}
& \multicolumn{3}{c}{\textbf{MIMIC-CXR}}
& \multicolumn{3}{c}{\textbf{IU-Xray}} \\
\cmidrule(lr){3-5}
\cmidrule(lr){6-8}
&
& \textbf{AUROC}
& \textbf{AUPRC}
& \textbf{F1}
& \textbf{AUROC}
& \textbf{AUPRC}
& \textbf{F1} \\
\midrule

\multirow{3}{*}{MedGemma}
& 5
& 79.68
& 90.12
& \textbf{84.08}
& 76.16
& 81.80
& 79.81 \\

& 10
& 79.61
& 90.10
& 84.06
& \textbf{76.55}
& \textbf{82.16}
& \textbf{79.95} \\

& 20
& \textbf{79.75}
& \textbf{90.23}
& 84.07
& 76.00
& 81.40
& 79.79 \\

\midrule

\multirow{3}{*}{LLaVA-Med}
& 5
& 78.79
& 94.73
& 90.98
& 78.45
& \textbf{89.50}
& 88.83 \\

& 10
& 79.01
& 94.72
& 91.00
& \textbf{78.56}
& 89.46
& \textbf{88.90} \\

& 20
& \textbf{79.12}
& \textbf{94.83}
& \textbf{91.22}
& 75.67
& 89.29
& 88.83 \\

\bottomrule
\end{tabular}
\end{table}



\subsection{Limitations and Future Work}

Although CLEAR consistently improves VIH detection across diverse medical VLMs, several limitations remain. First, CLEAR uses localized counterfactual intervention to probe phrase--image dependence rather than to establish formal causal attribution. The resulting response and alignment changes provide informative signals for VIH detection, while more causally grounded intervention strategies may further strengthen the interpretability of counterfactual auditing.

Second, the current framework focuses primarily on localized interventions in static two-dimensional medical images. Extending CLEAR to volumetric and temporal imaging modalities, such as CT, MRI, and longitudinal imaging studies, will require intervention strategies that preserve three-dimensional anatomical structure and temporal consistency.

Finally, CLEAR currently operates as a post-hoc VIH detection framework. An important future direction is to integrate phrase-level VIH risk into the generation process itself, enabling medical VLMs to dynamically verify and revise visually irrelevant content during generation. Such integration may extend CLEAR from post-hoc detection toward reliability-aware generation and self-corrective multimodal reasoning.

\section{Conclusion}
In this paper, we presented CLEAR, a model-agnostic post-hoc framework for phrase-level detection of visually irrelevant hallucinations in medical vision--language models. CLEAR operationalizes fine-grained visual-relevance auditing through phrase-conditioned grounding and localized counterfactual intervention. It then jointly models response and image--phrase alignment cues to estimate phrase-level VIH risk. Extensive experiments across radiology report generation and medical visual question answering benchmarks demonstrate consistent improvements over existing hallucination detection methods across both medical and general-purpose VLMs. CLEAR also remains robust across different perturbation operators and scales. In addition, we introduce a GT-derived controlled VIH benchmark, with hallucination labels adjudicated by consensus between two radiologists, to provide complementary evaluation under controlled clinically meaningful edits. Together, these findings support phrase-level counterfactual auditing as a practical approach for identifying visually irrelevant clinical content and provide a foundation for more reliable and self-corrective medical VLMs.

\bibliographystyle{IEEEtran}
\bibliography{main}

\newpage

\begin{IEEEbiographynophoto}{\bfseries Ke Zou}
received the Ph.D. degree in Computer Science from Sichuan University, Chengdu, China, in 2024. From 2022 to 2023, he was a Visiting Ph.D. Student with the Agency for Science, Technology and Research (A*STAR), Singapore. He is currently a Research Fellow with the Yong Loo Lin School of Medicine, National University of Singapore, Singapore. His research interests include medical image analysis, vision-language models, foundation models, and agentic AI. His work has been published in leading journals and conferences in medical imaging and artificial intelligence, including \textit{The Lancet Digital Health}, \textit{IEEE TPAMI}, \textit{IEEE TCYB}, \textit{MedIA}, and \textit{MICCAI}. He has served a Guest Editor for the \textit{IEEE JBHI} and served as an Area Chair for MICCAI 2025 and 2026.
\end{IEEEbiographynophoto}

\begin{IEEEbiographynophoto}{\bfseries Nan Zhou}
received the B.E. degree from Shanghai Ocean University, Shanghai, China, in 2022. She is currently pursuing the Ph.D. degree at Sichuan University, Chengdu, China. Her research interests include medical image analysis, deep learning, and uncertainty estimation.
\end{IEEEbiographynophoto}

\begin{IEEEbiographynophoto}{\bfseries Tao Yang}
received the B.E. degree from Sichuan Agricultural University, Ya'an, China, in 2025. He is currently pursuing the Ph.D. degree in computer science and technology at Sichuan University, Chengdu, China. His research interests include medical image analysis, multimodal large language models, and multi-agent systems, with a particular focus on their applications in healthcare.
\end{IEEEbiographynophoto}

\begin{IEEEbiographynophoto}{\bfseries Meng Liu}
is a Ph.D. candidate at Sichuan University. He received his Master degree from Sichuan University in 2023. He is now focusing on medical image analysis, multimodal learning and uncertainty estimation. 
\end{IEEEbiographynophoto}

\begin{IEEEbiographynophoto}{\bfseries Linchao He}
received the B.S. degree from Sichuan Agricultural University, Ya'an, China, in 2020. He is currently pursuing the Ph.D. degree at Sichuan University, Chengdu, China, where he is affiliated with the National Key Laboratory of Fundamental Science on Synthetic Vision. His research interests include medical image analysis, deep learning, and computational imaging inverse problems.
\end{IEEEbiographynophoto}

\begin{IEEEbiographynophoto}{\bfseries Zehui Liao}
is a Research Fellow at the Faculty of Information Technology, Monash University, Australia, where she works with Prof. Jianfei Cai. Before joining Monash University, she received her PhD in Computer Science from Northwestern Polytechnical University under the supervision of Prof. Yong Xia. From 2024 to 2025, she was a visiting PhD student at the Institute of High Performance Computing, Agency for Science, Technology and Research, Singapore, under the supervision of Dr. Huazhu Fu and Dr. Liangli Zhen. Her research interests include medical image analysis, with a particular focus on multimodal large language models and learning under label uncertainty.
\end{IEEEbiographynophoto}

\begin{IEEEbiographynophoto}{\bfseries Ling Huang}
received the Ph.D. degree in 2023 from Universit\'e de Technologie de Compi\`egne, Compi\`egne, France, where she conducted her doctoral research in a laboratory affiliated with the French National Centre for Scientific Research (CNRS). She is currently a Postdoctoral Researcher at Imperial College London, London, U.K., and a Visiting Research Fellow at the National University of Singapore, Singapore. Her research interests include the theory of belief functions and its applications to uncertainty quantification, causal inference, and decision explanation in medicine. She has authored or coauthored more than 40 papers, including over 20 papers published in leading journals and conferences, such as {Information Fusion}, {IEEE Transactions on Fuzzy Systems}, {IEEE Transactions on Cybernetics}, {Medical Image Analysis}, ICML, ECCV, ACL, ACM MM, and MICCAI. She serves as an Editorial Board Member for {Information Fusion}.
\end{IEEEbiographynophoto}

\begin{IEEEbiographynophoto}{\bfseries Guangren Huang}
is a Deputy Chief Physician in the Department of Radiology at Beihai People's Hospital, Guangxi, China. He is currently pursuing a Ph.D. degree at Guangxi Medical University. His research focuses on lung cancer, glioma, and medical imaging.
\end{IEEEbiographynophoto}

\begin{IEEEbiographynophoto}{\bfseries Yidi Chen}
received his Ph.D. degree from The First Affiliated Hospital of Guangxi Medical University and is currently an Associate Professor at the West China Hospital of Sichuan University. He was previously a joint doctoral candidate at the University of Wisconsin School of Medicine and Public Health. His research focuses on hepatocellular carcinoma and magnetic resonance imaging.
\end{IEEEbiographynophoto}

\begin{IEEEbiographynophoto}{\bfseries Yi Zhang}
received the Ph.D. degree in computer science and technology from Sichuan University, Chengdu, China, in 2012. He is currently a Full Professor with the School of Cyber Science and Engineering, Sichuan University. His research interests include medical imaging, compressive sensing, and deep learning. He serves as a Guest Editor for the {International Journal of Biomedical Imaging} and {Sensing and Imaging}, and as an Associate Editor for the {IEEE Transactions on Medical Imaging} and the {IEEE Transactions on Radiation and Plasma Medical Sciences}.
\end{IEEEbiographynophoto}

\begin{IEEEbiographynophoto}{\bfseries Hu Chen}
received the Ph.D. degree from the University of Electronic Science and Technology of China, Chengdu, China, in 2011. From 2008 to 2009, he was a visiting doctoral student at the University of California, San Diego, CA, USA. He is currently an Associate Professor with the School of Computer Science, Sichuan University, Chengdu, China. His research interests include image and video processing, computer vision, and artificial intelligence. He has authored or coauthored more than 40 papers in these areas.
\end{IEEEbiographynophoto}

\begin{IEEEbiographynophoto}{\bfseries Huazhu Fu}
(SM'18) is a Principal Scientist at the Institute of High Performance Computing (IHPC), A*STAR, Singapore. He earned his Ph.D. from Tianjin University in 2013. Prior to his current position, he was a Research Fellow at NTU, Singapore from 2013 to 2015, a Research Scientist at I2R, A*STAR, Singapore from 2015 to 2018, and a Senior Scientist at the Inception Institute of Artificial Intelligence, UAE from 2018 to 2021. His research focuses on medical image analysis, AI for healthcare, and trustworthy AI. He is an Associate Editor for several IEEE journals, including IEEE TMI, IEEE TNNLS, IEEE TAI, and IEEE JBHI.
\end{IEEEbiographynophoto}

\vfill

\end{document}